\documentclass[conference]{IEEEtran}
\IEEEoverridecommandlockouts

\usepackage{cite}
\usepackage{xcolor}
\usepackage{hyperref}
\usepackage{textcomp}
\usepackage{graphicx}
\usepackage{colortbl}
\usepackage{multirow}
\usepackage{booktabs}
\usepackage{subcaption}
\usepackage{algorithmic}
\usepackage{amsmath,amssymb,amsfonts}

\def\BibTeX{{\rm B\kern-.05em{\sc i\kern-.025em b}\kern-.08em
    T\kern-.1667em\lower.7ex\hbox{E}\kern-.125emX}}

\definecolor{red}{HTML}{fd8f8f}
\definecolor{greend}{HTML}{57e377}
\definecolor{greenl}{HTML}{b8fb8a}
\definecolor{yellow}{HTML}{fefdb4}
\definecolor{orange}{HTML}{ffd5ab}

\colorlet{red}{red!50}
\colorlet{yellow}{yellow!50}
\colorlet{greenl}{greenl!50}
\colorlet{greend}{greend!50}
\colorlet{orange}{orange!50}

\begin{document}

\title{Passage-Aware Structural Mapping \\ for RGB-D Visual SLAM}

\author{
    Ali Tourani$^{1}$, Miguel Fernandez-Cortizas$^{1}$, Saad Ejaz$^{1}$, David Pérez Saura$^{2}$, \\ Asier Bikandi-Noya$^{1}$, Jose Luis Sanchez-Lopez$^{1}$, and Holger Voos$^{1,3}$ \\
    \thanks{$^{1}$ Interdisciplinary Centre for Security, Reliability, and Trust (SnT), University of Luxembourg, Luxembourg.
    Holger Voos is also affiliated with the Faculty of Science, Technology, and Medicine at the University of Luxembourg, Luxembourg. \tt{\small{\{ali.tourani, miguel.fernandez, saad.ejaz, asier.bikandi, joseluis.sanchezlopez, holger.voos\}}@uni.lu}}
    \thanks{$^{2}$ Universidad Politécnica de Madrid, Spain. \tt{\small{david.perez.saura@upm.es}}}
    \thanks{$^{3}$ Faculty of Science, Technology, and Medicine at the University of Luxembourg, Luxembourg.}
}

% \author{
%     \IEEEauthorblockN{Ali Tourani}
%     \IEEEauthorblockA{\textit{SnT, University of Luxembourg}\\
%     Luxembourg City, Luxembourg \\
%     ali.tourani@uni.lu}
%     \and
%     \IEEEauthorblockN{Miguel Fernandez-Cortizas}
%     \IEEEauthorblockA{\textit{SnT, University of Luxembourg}\\
%     Luxembourg City, Luxembourg \\
%     miguel.fernandez@uni.lu}
%     \and
%     \IEEEauthorblockN{David Pérez Saura}
%     \IEEEauthorblockA{\textit{Universidad Politécnica de Madrid}\\
%     Madrid, Spain \\
%     david.perez.saura@upm.es}
%     \and
%     \IEEEauthorblockN{Asier Bikandi-Noya}
%     \IEEEauthorblockA{\textit{SnT, University of Luxembourg}\\
%     Luxembourg City, Luxembourg \\
%     asier.bikandi@uni.lu}
%     \and
%     \IEEEauthorblockN{Jose Luis Sanchez-Lopez}
%     \IEEEauthorblockA{\textit{SnT, University of Luxembourg}\\
%     Luxembourg City, Luxembourg \\
%     joseluis.sanchezlopez@uni.lu}
%     \and
%     \IEEEauthorblockN{Holger Voos}
%     \IEEEauthorblockA{\textit{SnT, University of Luxembourg}\\
%     Luxembourg City, Luxembourg \\
%     holger.voos@uni.lu}
% }

\maketitle

\begin{abstract}
Doorways and passages are critical structural elements for indoor robot navigation, yet they remain underexplored in modern Visual SLAM (VSLAM) frameworks. This paper presents a passage-aware structural mapping approach for RGB-D VSLAM that detects doors and traversable openings by jointly fusing geometric, semantic, and topological cues. Doors are modeled as planar entities embedded within walls and classified as traversable or non-traversable based on their coplanarity with the supporting wall. Passages are inferred through two complementary strategies: traversal evidence accumulated from camera–wall interactions across consecutive keyframes, and geometric opening validation based on discontinuities in the mapped wall geometry. The proposed method is integrated into vS-Graphs as a proof of concept, enriching its scene graph with passage-level abstractions and improving room connectivity modeling. Qualitative evaluations on indoor office sequences demonstrate reliable doorway detection, and the framework lays the foundation for exploiting these elements in BIM-informed VSLAM. The source code is publicly available at \href{https://github.com/snt-arg/visual_sgraphs/tree/doorway_integration}{https://github.com/snt-arg/visual\_sgraphs/tree/doorway\_integration}.
\end{abstract}


\section{Introduction \& Related Works}

% Intro VSLAM
Simultaneous Localization and Mapping (SLAM) has emerged as a fundamental capability of modern autonomous robots, allowing them to estimate their pose while incrementally reconstructing the surrounding environment \cite{slam_survey}.
Such a capability is crucial for robotic situational awareness \cite{slamtosa} and encompasses a broad range of tasks, including navigation, exploration, and inspection.
Among the available sensing modalities in SLAM, vision sensors provide a cost-effective means of capturing rich visual and structural data, leading to the emergence of Visual SLAM (VSLAM) \cite{vslam_survey}.
With the considerable progress in VSLAM, retrieving geometric information (\textit{e.g.,} 3D points) is often insufficient for understanding the environment, despite its impact on localization and basic mapping.
Accordingly, modern VSLAM approaches also incorporate the \textbf{semantic} and \textbf{structural information} of mapped entities, such as walls, chairs, and tables \cite{rso_slam, ps_slam}.

\begin{figure}[!t]
    \centerline{\includegraphics[width=\columnwidth]{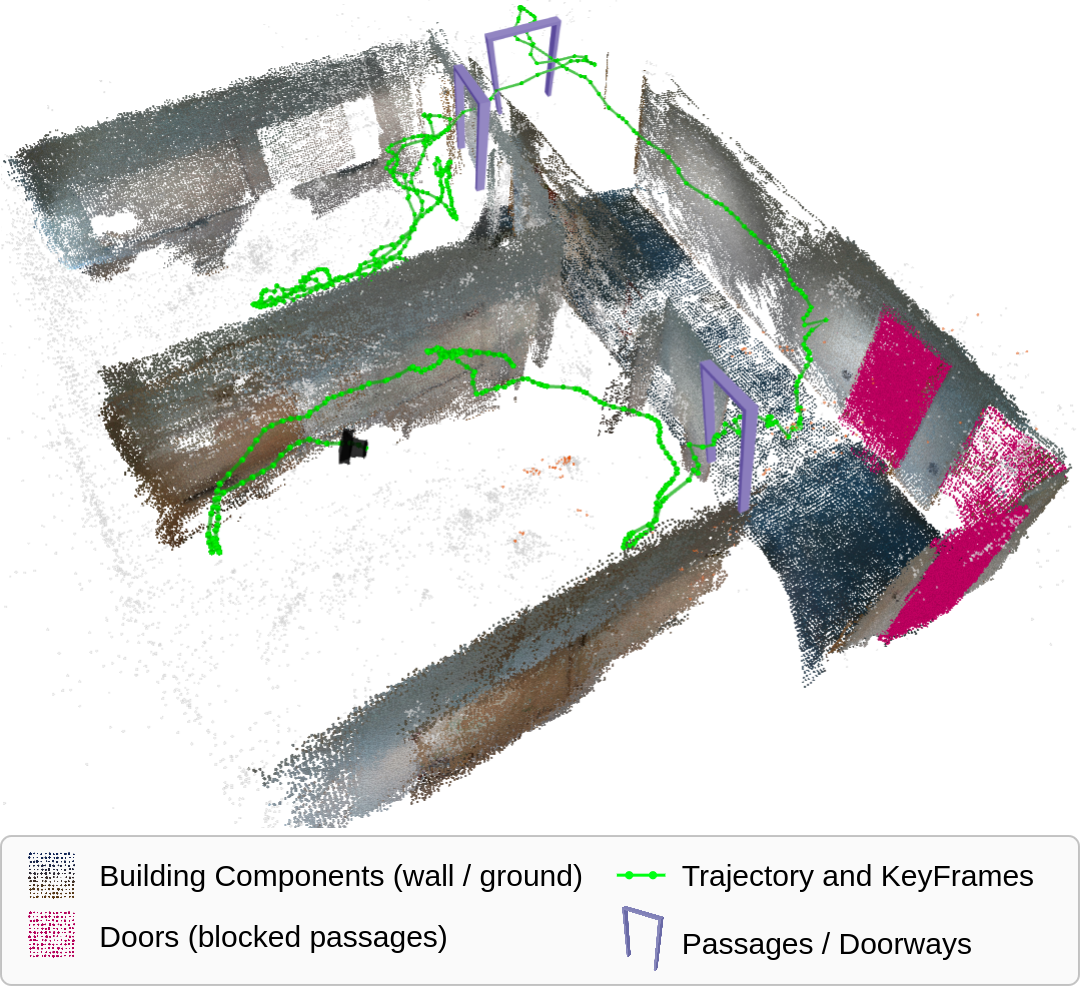}}
    \caption{Overview of door and passage mapping within vS-Graphs \cite{vsgraphs}, where semantic and geometric cues jointly localize traversable openings, enabling robust downstream tasks such as navigation and path planning.}
    \label{fig_overal}
\end{figure}

% Semantic VSLAM
Augmenting VSLAM with semantic information enables more interpretable and structurally meaningful map reconstruction.
In this direction, Tao \textit{et al.} \cite{ams_slam} integrate semantic perception and mapping to generate trajectories that maximize information gain with uncertainty reduction.
Despite the rich map reconstructions, the method introduces considerable system complexity and limited scalability.
SemanticFusion \cite{semanticfusion} produces temporally consistent semantic maps by probabilistically fusing predictions from multiple viewpoints at the expense of high computational cost.
Khronos \cite{khronos} employs semantic segmentation to identify short-term and long-term scene dynamics; however, its primary focus remains on dynamic objects rather than structural scene understanding.

Other approaches emphasize semantic representations of the environment layout, aiming to construct richer, more interpretable maps that include architectural entities such as walls and rooms.
vS-Graphs \cite{vsgraphs} proposes a tightly coupled framework that jointly optimizes VSLAM and 3D scene graph generation within a unified factor graph.
Incorporating \textit{building components} (\textit{i.e.,} walls and ground surfaces), \textit{structural elements} (\textit{i.e.,} rooms and floors), and their spatial relations produces geometrically grounded and semantically coherent environment representations.
However, the framework does not model \textit{doorways}, largely due to the absence of dense point cloud information.
In \cite{markervsg}, doorway detection and mapping are achieved using fiducial markers placed on door frames.
However, the approach relies on prior environmental instrumentation, which limits its practicality and scalability.

% Importance of Passages
It should be noted that passages are critical components of indoor spaces and have a special SLAM signature, as they define \textbf{traversable openings} between enclosed regions (\textit{e.g.,} rooms) and establish their inter-connectivity.
From a robotic perspective, passages are topologically meaningful landmarks that indicate transitions between navigable spaces.
Their detection extends room modeling beyond wall-only enclosures toward a richer structural-topological abstraction, supporting downstream tasks such as situationally aware navigation \cite{spath}.
Furthermore, incorporating prior knowledge from \textbf{Building Information Modeling (BIM)} can provide complementary structural constraints, enabling more reliable validation of passages.
Such priors enhance robustness in partially observed environments by aligning reconstructed geometry with known architectural layouts.
To address this gap, this paper makes the following contributions:
\begin{itemize}
    \item Formulating passage detection as traversable openings in walls by jointly fusing geometric, semantic, and topological cues.
    \item Integrating passages into a VSLAM system (vS-Graphs \cite{vsgraphs} as a proof of concept, extendable to other backbones), enabling improved room connectivity modeling and environment understanding.
    \item Publicly available source code to facilitate reproducibility and further research in the field.
\end{itemize}

\section{Proposed Method}

\subsection{Formalism}
\label{sec_formalism}

% Intro
Given an RGB-D point cloud at the VSLAM KeyFrame level, a panoptic segmentation method such as YOSO \cite{yoso} can be used to extract semantically meaningful planar entities, including walls and doors \cite{vsgraphs}.
Each KeyFrame is defined as \(K_t = \{\mathbf{P}_t, L_t, \varepsilon_t\},\) where $\mathbf{P}_t = \{\mathbf{p}_i \in \mathbb{R}^3\}$ denotes the 3D point cloud, $L_t$ the RGB image, and $\varepsilon_t$ the associated mapping metadata (e.g., camera pose).
Applying panoptic segmentation to $L_t$ yields pixel-wise semantic labels and instance-level masks, which are projected onto $\mathbf{P}_t$ to obtain semantically segmented point subsets \(\mathbf{P}_{\Psi}^{K_t} = S(\mathbf{P}_t, L_t) = \{\mathbf{P}_{wall}^{K_t}, \mathbf{P}_{door}^{K_t}, \dots\}, \) where each $\mathbf{P}_{\psi}^{K_t} \subset \mathbf{P}_t$ corresponds to a semantic class $\psi$.
To reduce redundancy and sensor noise, each segmented subset is refined through downsampling and range-based filtering, and then processed with RANSAC plane fitting \cite{ransac} to estimate semantically validated planar entities.
For each subset ${\mathbf{P}}_{\psi}^{K_t}$, a plane $\pi = (\mathbf{n}, d)$ is estimated, where $\mathbf{n} \in \mathbb{R}^3$ is the normal and $d \in \mathbb{R}$ the plane offset, such that a point $\mathbf{p}_r \in {\mathbf{P}}_{\psi}^{K_t}$ is considered an inlier if $\mathrm{dist}(\mathbf{p}_r,\pi) \leq \epsilon,$ with $\epsilon$ denoting the inlier threshold.
The validated planar entities detected in KeyFrame $K_t$ are subsequently inserted into the map and used for continuous structural reconstruction.
Among these semantic entities, locating \textbf{doors} and \textbf{passages} (\textit{e.g.,} doorways, archways, \textit{etc.}) is particularly valuable for downstream robotic tasks such as navigation and exploration.

\subsubsection{Doors}
\label{sec_formalism_door}

% Door
A door $\mathbf{P}_{door}^{K_t} \subset \mathbf{P}_{\Psi}^{K_t}$ is modeled as a \textit{physical planar structural object} extracted from RGB-D observations.
Since doors are typically embedded within walls, their \textit{traversability} can be inferred from their geometric relation to the associated supporting wall.
Let $\pi_d = (\mathbf{n}_d, d_d)$ and $\pi_w = (\mathbf{n}_w, d_w)$ be the estimated door and wall planes, respectively.
The door is ``\textit{closed/non-traversable}'' if it is approximately coplanar with its supporting wall:

\begin{equation}
    \cos^{-1}\left(|\mathbf{n}_d^\top \mathbf{n}_w|\right) < \tau_{\theta}
    \quad \text{and} \quad
    |d_d - d_w| < \tau_d.
\end{equation}

\noindent where $\tau_{\theta}$ and $\tau_d$ are predefined thresholds for angular alignment and coplanarity, respectively.
Satisfying both conditions indicates that the detected door lies approximately on the same plane as its supporting wall.

\subsubsection{Passages}
\label{sec_formalism_passage}

% Passage
A passage is defined as \(Q_i = \{\mathbf{P}_{wall}, \rho_i, \eta_i, m\}\), representing an opening (potentially blocked) embedded within a wall.
Here, $\rho_i \in \mathbb{R}^3$ denotes the passage centroid in the global reference frame, $\eta_i \in \mathbb{R}^2$ encodes its geometric extent (\textit{e.g.,} width and height), and $m$ stores its semantic variant (\textit{e.g.,} \texttt{doorway}, \texttt{archway}, or \texttt{unknown}).
When a closed door instance $\mathbf{P}_{door}$ is detected, a corresponding passage is instantiated at the same location and classified as a \texttt{doorway}, with its dimensions inferred from the associated door geometry. 
However, not all passages correspond to doors, and detecting generic traversable openings remains essential for tasks such as navigation and planning.
Accordingly, passage candidates are identified using two complementary strategies, as described below:

\begin{figure}[!t]
    \centerline{\includegraphics[width=\columnwidth]{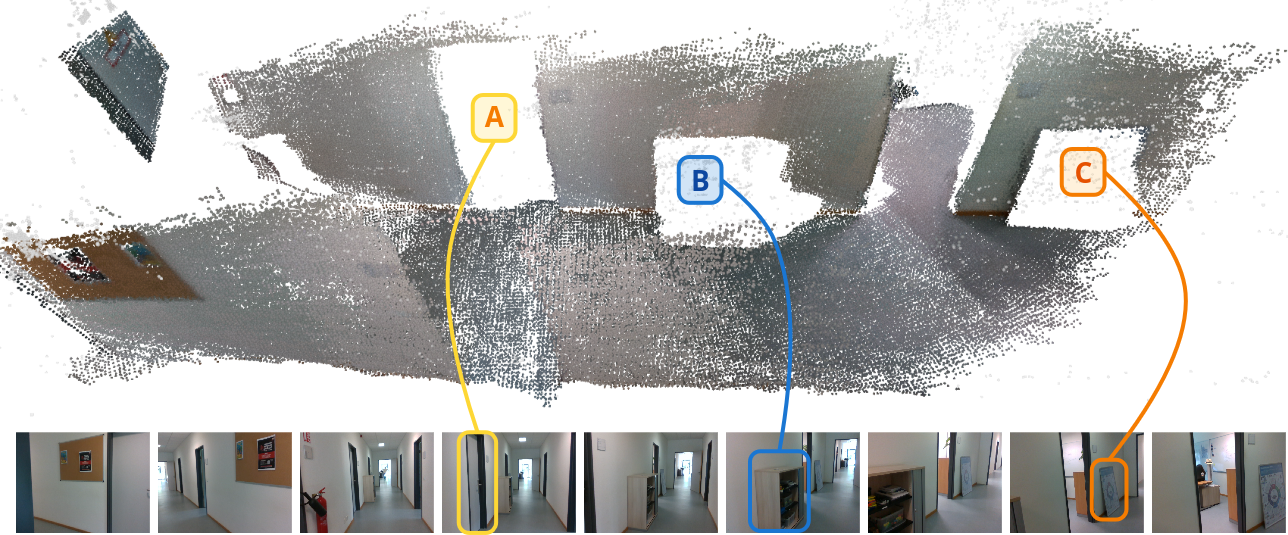}}
    \caption{Examples of geometric openings detected as gaps in semantic point clouds generated by vS-Graphs \cite{vsgraphs} across sequential frames, arising from (A) a door, (B) a storage cabinet, and (C) a poster on the ground, highlighting the ambiguity of purely geometric cues and the need for contextual validation.}
    \label{fig_possible_candidates}
\end{figure}

% Traversal Evidence
\noindent \textit{I. Traversal evidence-based approach}: passages are inferred from geometric and observational evidence across multiple KeyFrames by analyzing the interaction between the \textit{camera trajectory} and the \textit{mapped wall geometry}.
Having $\mathbf{c}_t$ as the camera center at KeyFrame $K_t$ and $\pi_w = (\mathbf{n}_w, d_w)$ as the supporting wall plane, the signed distance of $\mathbf{c}_t$ to the wall is $s_t = \mathbf{n}_w^\top \mathbf{c}_t + d_w.$
To ensure locality, only KeyFrames within a bounded distance to the wall ($d_{\max}$) are considered, \textit{i.e.,} $|s_t| < d_{\max}$.
A \textit{wall traversal event} is detected when two consecutive KeyFrames $K_t$ and $K_{t+1}$ exhibit a sign change in their signed distances, as $s_t \cdot s_{t+1} < 0$, indicating that the camera trajectory crosses the wall plane.
Such traversal events provide strong evidence of a \textit{local geometric opening} in the wall, as a true wall surface would otherwise prevent visibility and motion across it.
To improve robustness, traversal events are evaluated within a sliding temporal window over ordered KeyFrames, ensuring \textit{locally consistent} detections.
For each valid traversal event, the crossing point is estimated by linear interpolation between $\mathbf{c}_t$ and $\mathbf{c}_{t+1}$, yielding a candidate passage with centroid $\rho_i$.
If a door instance $\mathbf{P}_{door}$ is detected in proximity to a candidate passage centered at $\rho_i$, the passage is classified as a \texttt{doorway}, associated with the detected door, and its dimensions are inferred directly from the door geometry.
Otherwise, the passage remains \texttt{unknown}, and a default extent is assigned (\textit{e.g.,} $1.5\,\mathrm{m} \times 2\,\mathrm{m}$).

% Geometric Opening Analysis
\noindent \textit{II. Geometric opening validation approach}: in contrast to the local traversal-based strategy, passage candidates can also be inferred from a \textit{global analysis} of the reconstructed mapped wall geometry by explicitly searching for \textit{geometric discontinuities} or gaps.
Given a wall plane $\pi_w$ and its associated inlier point cloud $\mathbf{P}_{wall}$, the wall is projected onto a local 2D parameterization, where openings are identified as connected regions $\Omega_j \in \mathbb{R}^2$ with insufficient point support, \textit{i.e.,} $\forall \mathbf{u} \in \Omega_j,\; \mathcal{N}(\mathbf{u}) < \tau_\rho$, where $\mathcal{N}(\cdots)$ denotes local point density.
Each candidate region (\textit{i.e.,} gap) is then back-projected to the 3D reference, with a centroid $\rho_i$ and extent $\eta_i$.
Since such gaps may correspond to various entities (\textit{e.g.,} doors, windows, archways, or unmapped semantic objects near walls), as shown in Fig.~\ref{fig_possible_candidates}, a validation step is required to further filter out non-passage candidates.
This can be performed by evaluating \textit{geometric consistency} (\textit{e.g.,} size and aspect ratio of the gaps) and \textit{contextual cues} such as proximity to a detected door instance $\mathbf{P}_{door}$.
Only openings satisfying these constraints are retained as passages, while ambiguous regions are deferred until further evidence is accumulated across observations.

\subsection{Implementation \& Deployment}
\label{sec_implement}

\begin{figure}[!t]
    \centerline{\includegraphics[width=\columnwidth]{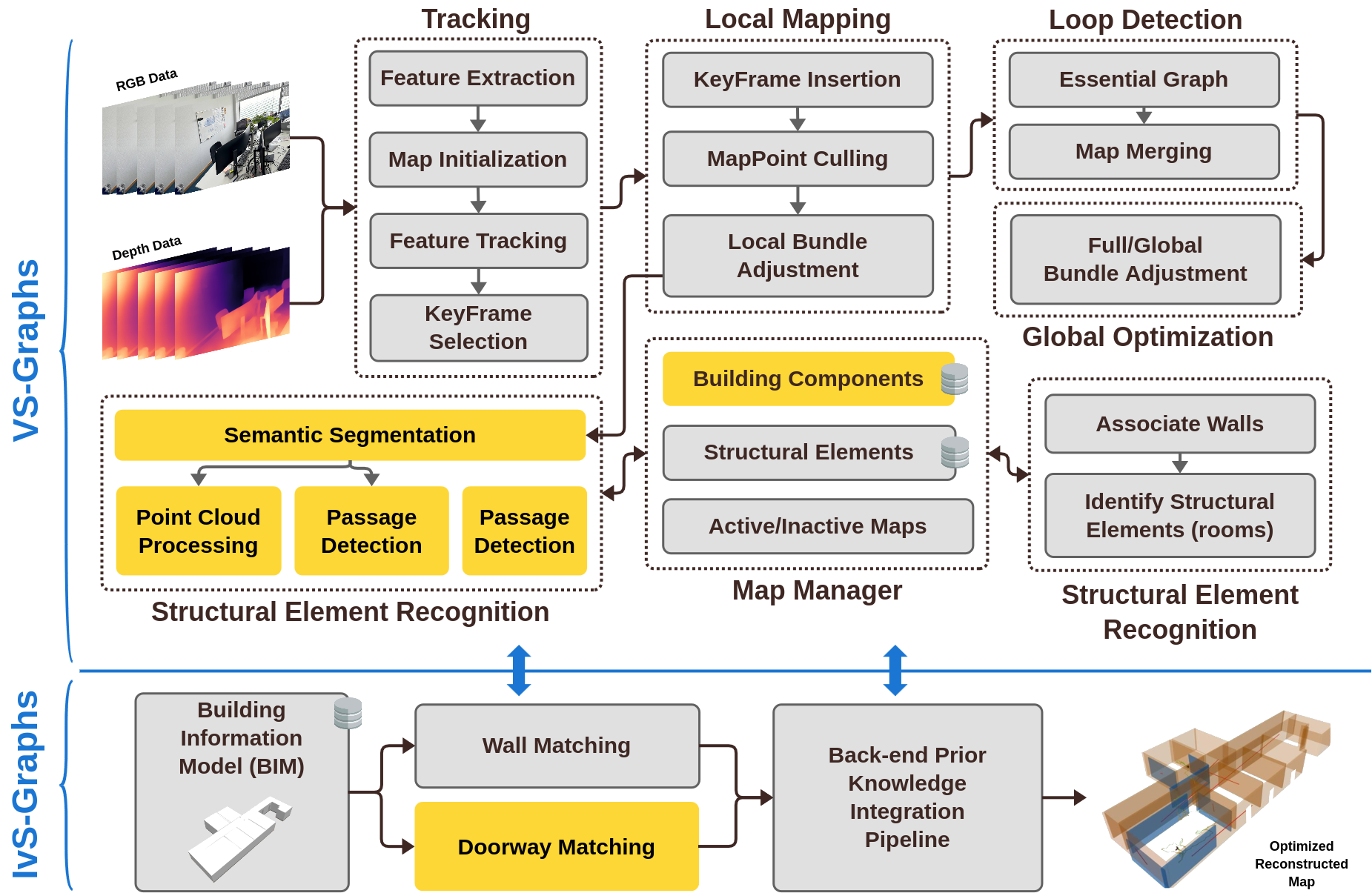}}
    \caption{System architecture showing the integration of passage detection within vS-Graphs \cite{vsgraphs} and its potential extension with structural priors in ivS-Graphs \cite{ivsgraphs} for improved scene understanding. Bordered, light gray modules indicate baselines' components, while highlighted modules denote the contributions of this paper.}
    \label{fig_solution}
\end{figure}

% Intro
The proposed passage detection framework has been integrated into vS-Graphs \cite{vsgraphs}, a real-time visual SLAM pipeline that enables online extraction of semantically verified structural elements.
This extension enables the extraction of additional \textit{geometric} and \textit{semantic cues} beyond those originally modeled in vS-Graphs (i.e., walls and ground surfaces).
In particular, it introduces higher-level abstractions that explicitly encode connectivity through passages within the environment.
Such enriched representations are beneficial for downstream tasks (\textit{e.g.,} navigation and planning) that rely on scene understanding (such as situationally aware path planning \cite{spath}), where reasoning about \textit{traversability} is essential.

% Integration vS-Graphs
Integration within vS-Graphs is depicted in Fig.~\ref{fig_solution}.
Accordingly, the \textit{building component recognition} thread is extended to support door detection by adding the door class to the semantic space (wall and ground).
This enables the identification of \textit{closed doors} within their surrounding wall context while mapping them.
Additionally, passage-related cues are detected in the \textit{structural element recognition} thread by periodically checking for openings in mapped walls.
This contains both \textit{traversal evidence} and \textit{geometric validation} approaches introduced in Section~\ref{sec_formalism_passage}.
It should be noted that, as these operations are performed at the KeyFrame level, the system preserves its real-time performance while enriching the scene graph with passage-level abstractions.

\begin{figure}[!t]
     \centering
     \begin{subfigure}[t]{0.38\columnwidth}
         \centering
         \includegraphics[width=\textwidth]{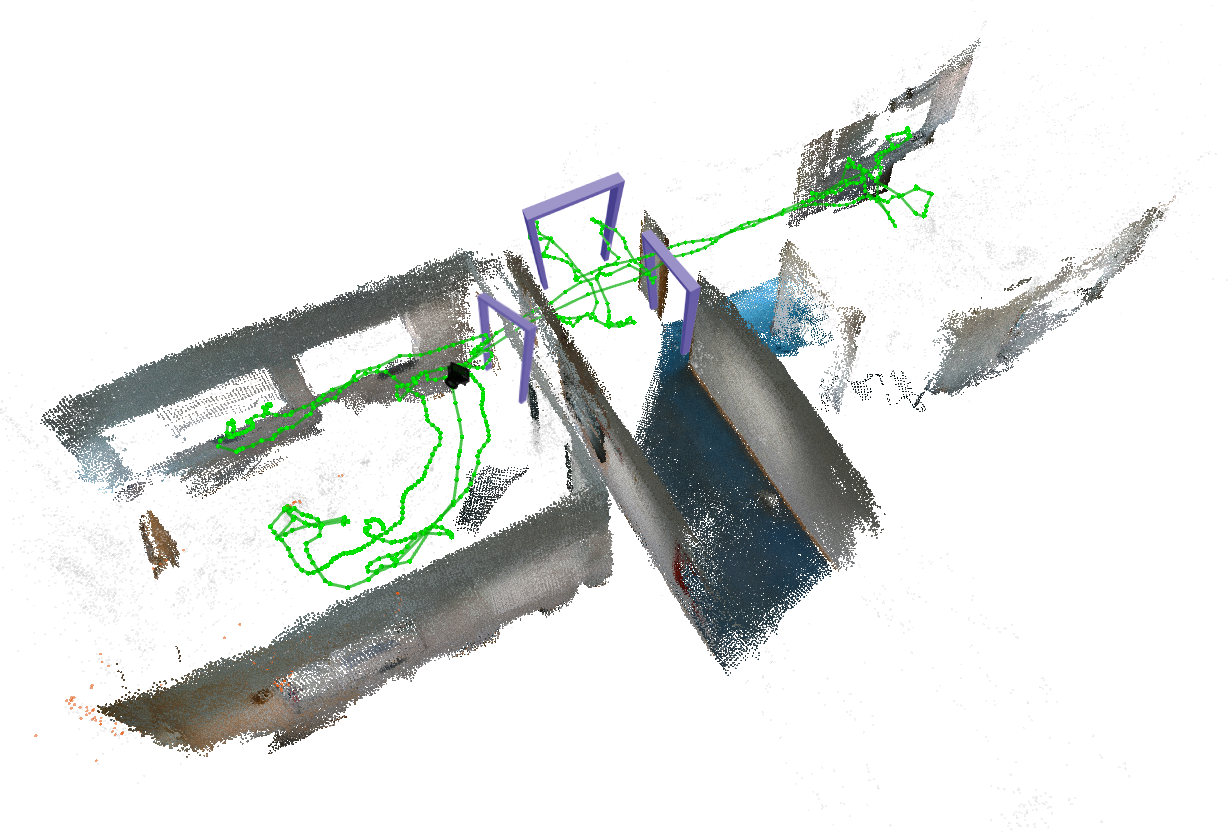}
         \caption{SMapper - Seq. MR01}
     \end{subfigure}
     \begin{subfigure}[t]{0.6\columnwidth}
         \centering
         \includegraphics[width=\textwidth]{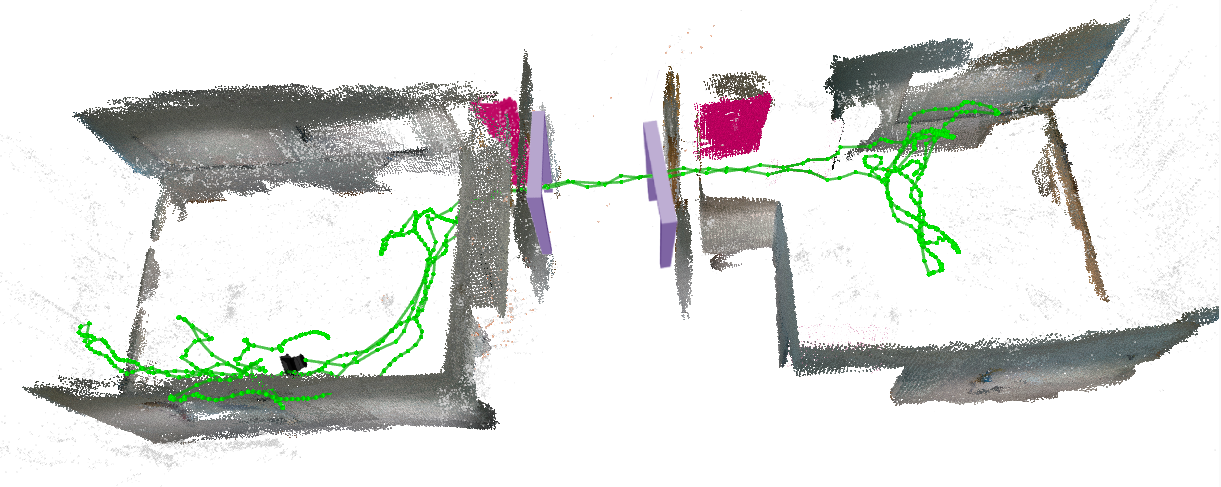}
         \caption{SMapper - Seq. MR04}
     \end{subfigure}
     \caption{Qualitative results of the proposed passage detection and integration approach in indoor office environments, with vS-Graphs \cite{vsgraphs} as the baseline.}
     \label{fig_qualitative}
\end{figure}

\section{Benchmarking \& Evaluations}

% Intro
The proposed approach has been evaluated through qualitative analyses of dataset instances recorded in indoor office environments with the SMapper device \cite{smapper}.
The collected data is used to validate the effectiveness of the proposed passage detection pipeline.
Experimental results demonstrate that the system reliably identifies passages, primarily doorways, using both \textit{traversal evidence} and \textit{geometric validation} approaches, as illustrated in Fig. \ref{fig_qualitative}.
In the figure, light purple doorway markers indicate open passages, while crimson point clouds correspond to closed or obstructed doorways.
The results further highlight the consistency of the proposed method across varying spatial layouts and levels of environmental complexity.

\section{Potentials \& Discussions}

\begin{figure}[!t]
    \centerline{\includegraphics[width=0.75\columnwidth]{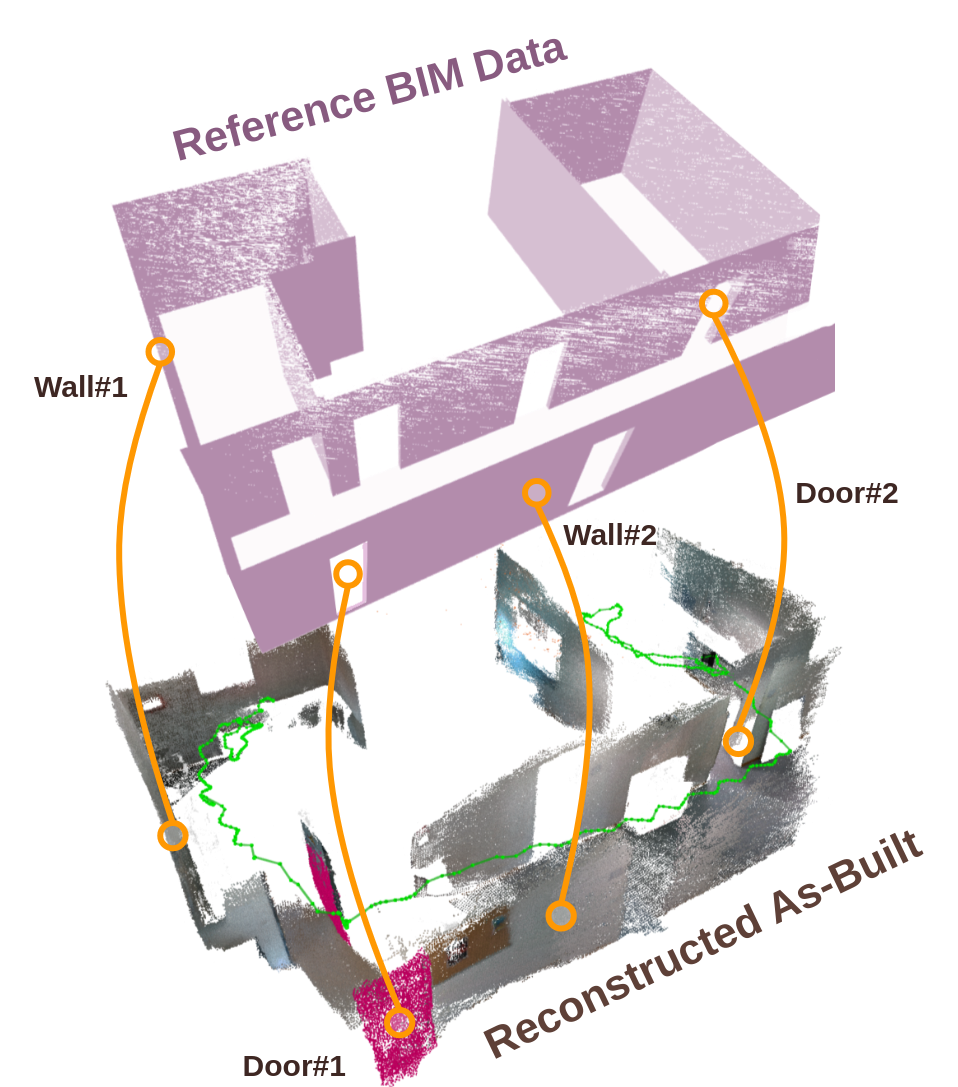}}
    \caption{Potential integration of the proposed passage detection methodology within the structural priors in ivS-Graphs \cite{ivsgraphs} framework.}
    \label{fig_potential}
\end{figure}

% ivS-Graphs
The proposed approach can be augmented with a BIM-informed extension of vS-Graphs, as introduced in \cite{ivsgraphs} (titled ivS-Graphs), thereby incorporating prior structural knowledge into the mapping process.
Such an extension enables the establishment of correspondences between the reconstructed scene and the as-planned BIM model, allowing detected passages to be more robustly validated and refined.
This integration improves scene understanding by reducing geometric ambiguity and increasing the structural consistency and completeness of the reconstructed environment.
As depicted in Fig.~\ref{fig_solution}, the underlying vS-Graphs framework is naturally coupled with BIM information, facilitating more robust reasoning about structural elements.
In particular, the availability of prior knowledge simplifies the validation of doorway existence and enhances confidence in passage detection, as further depicted in Fig.~\ref{fig_potential}.
Accordingly, the current version of ivS-Graphs enables the detection of structural deviations by establishing correspondences between \textit{as-built} walls in the SLAM reconstruction and their \textit{as-planned} counterparts.
This has been effectively demonstrated for wall alignment, where inconsistencies can be identified and corrected using BIM priors.
Building on this capability, the proposed framework extends this principle to \textit{passages}, enabling detected doorways and openings to be validated against the BIM model.
This facilitates the identification of missing, misaligned, or falsely detected passages, further improving the structural reliability of the reconstructed environment.

% Direction Path Planning
Additionally, and from an application perspective, enriching reconstructed maps with passages significantly enhances their utility for downstream robotic tasks.
In particular, explicitly modeling doorways and traversable passages provides critical information for path planning and navigation, enabling robots to reason about connectivity between spaces.
Such representations improve situational awareness by allowing more informed decision-making, especially in complex indoor environments where accessibility and transitions between regions are essential.

\section{Conclusions}

This paper introduced a passage-aware structural mapping approach for RGB-D Visual SLAM, formulating doorway detection as the identification of traversable openings in walls through the joint use of geometric, semantic, and topological cues. The method has been integrated into vS-Graphs and qualitatively evaluated on indoor office sequences, where it reliably distinguishes between open passages and closed or obstructed doorways while preserving real-time operation at the KeyFrame level. Beyond the current proof of concept, the framework is designed to be naturally extended through BIM priors, as outlined in ivS-Graphs\cite{ivsgraphs}, to validate detected passages against as-planned models and improve structural consistency. Future work will focus on incorporating doors and doorways directly within the factor graph optimization, enabling tighter coupling between traversability reasoning and pose estimation, as well as quantitative benchmarking across more diverse indoor environments.

% \section*{Acknowledgment}
% Sample

\end{document}